# Choosing features for classifying multiword expressions

Eric Laporte
Université Paris-Est, Laboratoire d'informatique Gaspard-Monge CNRS UMR 8049, F77454 Marne-la-Vallée, France

*Multiword expressions (MWEs) are a heterogeneous set with a glaring need for classifications. Designing a satisfactory classification involves choosing features. In the case of MWEs, many features are a priori available. Not all features are equal in terms of how reliably MWEs can be assigned to classes. Accordingly, resulting classifications may be more or less fruitful for computational use. I outline an enhanced classification. In order to increase its suitability for many languages, I use previous works taking into account various languages.*

## 1 Introduction

Multiword expressions range from idioms like *put pen to paper*, meaning 'undertake to write something', to multiword terms like *protein kinase* to support-verb constructions like *take a dip* 'bathe' and other types. Due to such diversity, there is a glaring need for classifications, if only for practical organization and for necessities of communication within the research community. Forty years after the first published comprehensive classifications of sets of MWEs, the community has not reached a satisfactory consensus on large classes or on the most relevant features. One outline of a classification (Sag *et al.* 2002), based on Nunberg *et al*. (1994), is influential, but some classes are fuzzily defined. The community is seeking to delineate the basic objects of the field. This uncertainty confuses computer scientists' main MWE-related activity, which is to recognise types of MWEs in texts through statistical engineering: the community does not offer them a consensual definition of types of MWEs.[1]

Classifications are a matter of features of the items to be classified. Which features should be used for classification, and therefore investigated in priority? Of course, linguistic relevance plays a prominent role in this selection, but my point in this paper is that many researchers overlook other important reasons for selecting or discarding some kinds of features. Some features are fuzzy and imprecise, that is, it is difficult to tell which MWEs have them. In resulting classifications, assignment of MWEs to classes is less reliable than it could be, and this is detrimental to computational use. Other features are more clear-cut and potentially more useful, but have not made their way to computational-linguistic literature yet. Another requirement for a convenient classification is that its outline be suitable for many languages. Accordingly, I use previous work taking into account various languages.

In §2, I exemplify and discuss the notion of a fuzzy feature. In §3 and §4, I investigate two connected topics: clusters of correlated features, and practical problems of observation. §5 advocates in favour of the practice of checking information against the lexicon. I outline an enhanced classification in §6.

[1] However, there is a relative consensus on the delimitation of MWEs themselves. At least, many experts agree tthat this class includes collocations, multiword terms and support verb constructions. I will not address this issue in more detail for lack of space.

# 2 Clear-cut or fuzzy features?

## 2.1 Examples

Some features are more clear-cut than others. For example, some MWEs select a preposition for a free slot/argument position,[2] as in *have pity on*:

(1) *You could have pity on us*

Nothing is totally definite in linguistics, but using *on* in this context is clearly appropriate.

In contrast, the semantic weight of verbs is a much fuzzier feature that lies in a continuum. The verb *have* in (1) is deemed "light", whereas it has full semantic weight in:

(2) *They will have this machine in soon*
‘This machine will soon be available for sale in their store’

This is a basis to classify *have pity* in (1) as a support-verb construction, or light-verb construction, and *have in* in (2) as a phrasal verb. But, in *have a call* ‘talk on the phone’, or *have a goal,* or *make a joke,* intuition about the semantic weight of the verb in these expressions remains unsettled or depends on whom you ask.[3]

## 2.2 Related work

### 2.2.1 Earlier work on clear-cut features

All the main features for present classifications had already been proposed by 1995, so the historical background is worth reviewing.

The first research works on MWEs with extended classificatory results define classes and subclasses with relatively clear-cut features. For instance, Labelle (1974)’s study of French support-verb constructions with *avoir* ‘have’ assigns a class to expressions with an argument position introduced by the preposition *sur* ‘on’, as in:

(3) *Lyon a un avantage sur Marseille*
‘Lyon has an advantage over Marseille’[4]

This kind of sharp distinction neatly separates classes. For example, *avoir un faible pour* ‘have a taste for’ definitely does not select *sur*, since sequences like (4) are rejected:[5]

(4) **J’ai un faible sur toi*
‘*I have a taste on you’

The other features used to define Labelle’s classes are similar. Many features come down to applying elementary syntactic operations, one at a time, and judging the acceptability of the result, while watching out for unexpected meaning changes. The method used by Labelle, called Lexicon-Grammar (LG) by Guillet & La Fauci (1984), is briefly described by Gross (1994). It was applied to MWEs by other authors since then (Meunier 1977; Giry-Schneider

[2] Here, *free* means that the content of the slot, i.e. the noun phrase, is variable.

[3] Another fuzzy feature of MWEs is whether they belong to terminology. *Protein kinase* does, *smooth operator* ‘persuasive person; manipulative person’ does not; but *sore throat* ‘inflammation of the throat’ is somewhere in between, since it is used by professionals but mainly to communicate with non-professionals.

[4] For examples not in English, I do not provide glosses because they would not be useful for the reader. I provide a translation of the literal meaning when it is different from the non-literal meaning.

[5] Independently of that, *avoir un avantage* may also occur with other prepositions, maybe less clearly selected:

*J’ai un avantage par rapport à toi*
‘I have an advantage as compared to you’

1978; Danlos 1980; Gross 1982; Freckleton 1985; Machonis 1985; Ranchhod 1990; etc., in English, Romance languages, Greek, etc.). All prefer clear-cut features such as:

• parts of speech (multiword nouns; verbal, adverbial and adjectival idioms),

• applicable syntactic operations, including optionality vs. compulsoriness of fixed constituents and free slots.[6]

Some examples of clear-cut features are less likely to occur at the top of a classification tree:

• phrase structure (e.g. number of fixed objects in a verbal idiom),

• number of free slots, their selected prepositions, restrictions on what may fill them,

• compulsory coreference relations (e.g. in *think on one's feet* 'improvise a reaction quickly', between the free subject and the possessive).

Nothing is totally definite in linguistics, but the implicit rationale behind preference for clear-cut features is that it is unwise to place poorly understood features in a decision tree, especially at its top.

### 2.2.2 Earlier work on fuzzy features

However, outside this LG trend, clear-cut features are readily mixed with fuzzier ones, even when defining large classes. The clear-cut features are essentially the same as above. The fuzzier ones often involve semantics or psycholinguistics.[7]

SEMANTIC WEIGHT is often used to define support-verb constructions, or light-verb constructions, such as *have pity, have a goal, take a dip*. This section will show that this definition relies entirely on fuzzy features, and so does the other naive definition.

Baldwin & Kim (2010: 276) define light-verb constructions by the fact that their verb is "semantically bleached or 'light', in the sense that [its] contribution to the meaning of the light-verb construction is relatively small in comparison with that of the noun complement," that is semantically weak. This definition dates back to Jespersen: "such everyday combinations as those illustrated in the following paragraphs after *have* and similar 'light' verbs (...) are in accordance with the general tendency of modern English to place an insignificant verb, to which the marks of person and tense are attached, before the really important idea (…) *I really must have a good stare at her*" (Jespersen 1942: 117). But in many occurrences, verbs are felt to lie somewhere in a spectrum of intermediate stages between significant and insignificant. And, even though the feature is polar, it is not scalar: there is no metrics according to which it would be possible to measure how close an item is to the poles of the range.

Alternative views of support-verb constructions have been proposed. One of them is in terms of predicate-argument structure: "nouns that have characteristics of predicates" (Gross 1981: 32, my translation; Cattell 1984). By *predicate-argument structure*, I mean the concept borrowed from logic by linguists, who initially applied it (Tesnière 1959) to sentences such as

(5) *The wire connects the device to the socket*

[6] Fraser (1970: 39) also proposes a classification of verbal idioms based on applicable syntactic operations. But he tests it on a sample of 131 idioms only (p. 40-41). In addition, he hypothesizes entailments between operations: for instance, if an idiom accepts passivization, it would also accept permutation of complements. His classification presupposes the entailments: when some of them are wrong for an idiom, there is no class for it. This is the case for the French idioms *faire le jour sur* (lit. make the day on) 'shed light on' and *claquer la porte au nez de* (lit. slam the door to the nose of) 'slam the door in the face of', which accept passivization, but not permutation of complements.

[7] Wisely, current classifications of MWEs avoid using terminologicalness to define main classes.

In this analysis, the predicate-argument structure of (5) is ‘connect’(‘wire’, ‘device’, ‘socket’), where the predicate is ‘connect’ and the arguments are ‘wire’, ‘device’ and ‘socket’. The predicate does not necessarily match with a verb:

(6) *Everyone took a look at our project*

Analysing (6) as ‘take’(‘everyone’, ‘look’) or ‘take’(‘everyone’, ‘look’, ‘project’) is not satisfactory, precisely because *take* is too weak to make sense as the core of a predicate-argument structure. If you analyse (6) as ‘take_look’(‘everyone’, ‘project’) instead, you consider that the predicate is *take a look* (or the noun *look*, which makes little difference, since *take* has features of a mere function word). Or, in other words, the noun *look* has valency two. On the basis of this type of analysis, support-verb constructions could be defined as those in which the predicate does not match with the main verb, but with a noun (a predicational noun, or noun that has valency) or another part of speech (PoS). Unfortunately, this definition still relies on a shaky semantic intuition: which part of a sentence matches best with the intuition of predicate? Take the following sentence:

(7) *He made a joke*

In (7), is the verb *make* to be analysed as a “performance” predicate, or is it so light that the sentence is equivalent to *He joked*?

Another alternative is suggested by examples from Jespersen (1942), which all involve deverbal nouns such as *stare*, and by the pairs of sentences explicitly pointed out by Harris (1964: 17–19):

(8) a. *He took a look at it*
b. *He looked at it*

Could support-verb constructions be defined by the equivalence between the content verb (*looked*) and the support-verb followed by the deverbal noun (*took a look*)? This would be consistent with both previous definitions.[8] Unfortunately, the definition based on equivalence with a verb would exclude many expressions for which no equivalent verb is in use (Labelle 1974):

(9) a. *Il a eu un conflit avec sa famille*
‘He had a conflict with his family’

b. **Il s’est conflité avec sa famille*
(lit. He conflicted himself with his family)
‘He conflicted with his family’

This is not desirable because (9a) otherwise behaves like a typical support-verb construction. It is syntactically and semantically similar, for example, to (10a), for which an equivalent verb is observed:

(10) a. *Il a eu une réconciliation avec sa famille*
‘He had a reconciliation with his family’

b. *Il s’est réconcilié avec sa famille*
(lit. He reconciled himself with his family)
‘He was reconciled with his family’

Here are parallel examples in English:

[8] As for the definition based on semantic weight, if *look* is equivalent to *take a look*, little is left for *take* to contribute to the meaning. Now for the definition referring to predicate-argument structure: if *looked* is the predicate in (8b), its equivalent *took a look* should logically be considered as the predicate in (8a).

(11) a. *He has the goal of getting rich*
b. **He goals to get rich*

(12) a. *He has the aim of getting rich*
b. *He aims to get rich*

Reformulating, the property of equivalence with a content verb would not classify (9a) and (11a) as support-verb constructions, in spite of their striking similarity with (10a) and (12a).

Thus, we are left with the first two naive definitions of support-verb constructions: one with the semantic weight of the verb, and the other with predicate-argument structure. Both definitions rely on particularly fuzzy semantic intuitions. They situate the feature of being a support-verb construction in a continuum between two poles. A more precise definition will be reported in §2.2.3.

Gibbs & Nayak (1989: 104) define another loose feature, SEMANTIC DECOMPOSABILITY, as the "[contribution of] parts of idioms to their figurative interpretations as a whole [according to] speakers' assumptions". For example, the parts of *pull strings* 'covertly use one's influence on personal connections' would be *pull* 'exploit' and *strings* 'personal connections'. This is a continuously graded intuition: "People's intuitions about the decomposability of any idiom can vary along some continuum of semantic decomposition" (Gibbs & Nayak 1989: 67); "in general, idiom phrases exist on a continuum of analyzability ranging from those idioms that appear to be highly decomposable (e.g., *pop the question*) to those that can be viewed as semantically nondecomposable (e.g., *kick the bucket*)" (p. 107). Nunberg *et al.* (1994: 508) reterm this feature SEMANTIC ANALYSABILITY and redefine it as the fact that the "idiomatic interpretation [can] be distributed over [the] parts of the [expression]" (p. 497). The wording is different, but it comes to the same thing, since the only source to know the distribution of the idiomatic interpretation over the parts of the expression is speakers' assumptions.[9] Nunberg *et al.* cite an uncertain case: they represent *take advantage of* with two lexical entries, one of which is semantically analysable while the other is not, although they "know of no evidence that the two entries might be semantically different" (pp. 520–523). This case where the same idiom, in the same sense, both is and is not analysable implicitly situates it at some intermediate stage. Although analysability is imprecise, Sag *et al.* (2002) and Baldwin & Kim (2010: 270) adopt this feature, going back to the term of SEMANTIC DECOMPOSABILITY,[10] to distinguish two of their major classes of MWEs: semi-fixed expressions and syntactically-flexible expressions.

Thus, reputed classifications use fuzzy features as liberally as clear-cut ones, even at the top of their classification trees.

### 2.2.3 Clear-cut features and lexical inventorying

There is something more to be learned from early work on MWEs: extensive practice of lexical description leads researchers to discover clear-cut features and adopt them in their classifications.

Recall that Labelle (1974) and other LG authors cited in §2.2.1 prefer clear-cut features. This specificity is connected to their practice of inventorying lexical items: they delimit a set of phrases on the basis of features, systematically record phrases belonging to the set, obtain comprehensive lists and study them in order to reach well-documented conclusions. The papers and PhDs of these linguists either include a comprehensive list of members of each class proposed, or at least were published after the completion of such lists.[11] For example, Freckleton (1985) lists 8000 English verbal idioms; by 1987, Gross' laboratory[12] had studied 12700 entries of French predicational nouns used with support verbs (Tolone 2011: 144).

[9] Nunberg *et al.* (1994: 496–497) contrast semantic analysability with *transparency*, which is about speakers' ability to guess why an expression with some literal meaning is used to convey a given non-literal meaning.

[10] Baldwin & Kim (2010: 270) equate their notion of decomposability to Nunberg *et al.* (1994: 496)'s semantic analysability.

This labour-intensive method contrasts with common practice of that time. Nunberg *et al.* (1994), for example, do not challenge the LG approach or the resulting classifications,[13] but base most of their research on sporadically picked examples: the only sizable lists reproduced in their paper (pp. 532–534) answer empirically one of the many issues they address. When they claim that the number of MWEs with anomalous morphosyntactic structure, like *every which way* 'in all directions; in complete disorder', is "not so small" (p. 515), as a reply to Chomsky (1980: 149) who claims such expressions are not "typical", they do not compare numbers of lexical entries in comprehensive dictionaries. (Tables of MWEs settle this dispute in favour of Chomsky: morphosyntactically anomalous MWEs are really a small minority.)

Sag *et al.* (2002) and Baldwin & Kim (2010) share the same weaknesses. In no language did any research group assign semantic (un)analysability to comprehensive classes of MWEs. Bond *et al.* (2015: 64) encode this property in 421 English idioms, but this is a small sample, not a comprehensive lexical inventory; in contrast, Grégoire's study of 5000 Dutch MWEs led her to give up categorizing them as analysable or not (Grégoire 2010: 31-32).

But what is the connection between clear-cut features and extensive lexical description? When the description of a feature gives clear-cut results throughout the inventory of expressions, authors understandably tend to consider these results particularly reliable, and to prefer this feature over others, all else being equal.

This is how Gross and his followers in the 1970s came across formal features of support-verb constructions which are still used as criteria to recognise them (Langer 2005). They systematically scanned the lexicon of French nouns, studied their syntactic constructions and worked out criteria of recognition of predicational nouns for dubious cases. One of these criteria is a formal property of determiners and adjuncts (Gross 1976: 109) which is also observed in English: in (13a), if possessive determiners, phrases with *of* and genitives are inserted around *joke*, they cannot refer to anything else than the subject:

(13) a. *He made a joke*
b. *He made his joke*
c. **He made your joke*
d. **He made Ann's joke*

How does this criterion correlate with the intuitive notion of predicate (cf. §2.2.2)? In other sentences where the core of the intuition-identified predicate is a noun, like (8a) *He took a look at it* or (11a) *He has the goal of getting rich*, this constraint is also observed. But when the intuitive predicate is a verb, the constraint is not observed:

(14) a. *He made your car*
b. *He made Ann's car*

[11] The lists still exist. They describe features for all entries and take the form of tables of features, which are easy to use. Many of these tables are freely available, e.g. at http://infolingu.univ-mlv.fr/ for French. They remain to be diffused so that they reach out to the mainstream community.

[12] Laboratoire d'automatique documentaire et linguistique (LADL), a part of Université Paris 7 and of CNRS.

[13] Nunberg *et al.* (1994: 498)'s divergence from Machonis (1985) is terminological: they call *conventionality* what Machonis (1985: 306) and Danlos & Gross (1988: 128-129) call *lack of compositionality*, that is impossibility of predicting the meaning or use of the MWE on the basis of *only* a knowledge of the rules that determine the meaning and use of its parts when they occur separately.

(15) a. *He reported your joke*
b. *He reported Ann's joke*

This formal test is a reason for analysing (13a) as 'joke'('he'), but (14a) as 'make'('he', 'car') and (15b) as 'report'('he', 'joke'('Ann')), in spite of their apparently similar structure.

This property, when used as a criterion to distinguish support-verb constructions from full-verb constructions, gives more precise results than those I mentioned in §2.2.2. It does not help with the distinction between support-verb constructions and verbal idioms. Some verbal idioms behave as (13):

(16) a. *He thought on his feet*
'He improvised a reaction quickly'

b. **He thought on Ann's feet*

Others behave as (14) and (15):

(17) a. *He melted your heart*
'He made you feel sympathy'

b. *He melted Ann's heart*

But a similar property (Gross 1979: 865–866, footnote 6), which is used in a forthcoming compendium on French grammar (Abeillé & Vivès 2011: 16), contributes to making more definite the distinction between support-verb constructions and verbal idioms. Take the following construction with the support verb *make*:

(18) *The quake made damage to the area*

A syntactic operation applied to (19a) produces a variant (19b) where *make* is absent:

(19) a. *The damage to the area made by the quake (is described in the diary)*
b. *The damage to the area by the quake (is described in the diary)*

This criterion classifies *have the back of*, meaning 'back, support', as a verbal idiom, not as a support-verb construction, because it has no variant in which *have* would be absent. Take the following sentence:

(20) *The president has the back of our children*

A banal syntactic operation on (20) would produce the subject of the following sentence:

(21) **The president's back of our children (is manifested by real actions)*

But (21) is not in use.

This criterion applies to all support-verb constructions. The syntactic operations that remove the support verb are not the same for all support-verb constructions, even in a given language: (19) exemplifies one of them for English, (20)-(21) another. Applying the criterion may involve knowing all operations and testing them, because the criterion rules out the support-verb construction analysis only if none applies, e.g. for *have the back of*.[14] In Italian (De

[14] This criterion also rules out the support-verb construction analysis for *take advantage of* in: (i) *The bank takes advantage of deposit slip errors*. The syntactic operation of (19) does not apply:

(ii) *The advantage taken by the bank of deposit slip errors (has been revealed)*
(iii) **The advantage by the bank of deposit slip errors (has been revealed)*

Neither does that of (20)-(21):

(iv) *The advantage the bank takes of deposit slip errors (has been revealed)*
(v) **The bank's advantage of deposit slip errors (has been revealed)*

Angelis 1989), Portuguese (Ranchhod 1990, Rassi *et al.* 2014), Korean (Han 2000), Greek (Kyriacopoulou & Sfetsiou 2003) and other languages, LG authors selected analogous technical criteria and definitions.

The larger linguistic and NLP community was not receptive to LG in the 1980s and 1990s, and access to publications was difficult. Cattell (1984), though he did not explicitly challenge the LG syntactic criteria, stuck to definitions based on semantic intuition, and so did many linguists. Since then, both traditions, that is intuitive vs. technical definitions of support-verb constructions, have continued in parallel.[15] Thus, support-verb constructions provide an example of a feature that can be defined either as clear-cut or as fuzzy. The notion itself is basically the same, but some definitions ensure more definite membership than others.

This review of earlier work showed that researchers engaged in projects of extensive lexical description tend to discover more clear-cut features and adopt them in their classifications, but that recent literature does not make a difference between clear-cut and fuzzy features. Recent classifications are derived indifferently from both.

## 2.3 Discussion

I could not find in the literature any discussion of whether the use of these fuzzy features is an issue at all, or if their relevance compensates for their drawbacks. Even NLP literature does not care more than corpus linguistics or the philological or generative traditions.

Fuzzy features can be less technical. For example, as opposed to the definition of support-verb constructions based on the semantic weight of the verb, a concern for definiteness leads one to adopt formal criteria that involve applying detailed syntactic operations, as in (13c) **He made your joke*, and assessing the result. It is not just that this complexity can be seen as a drawback: it also makes precise features more likely to be language-dependent. The criteria illustrated by (13) for English need to be adapted when they are applied to, say, Italian or Korean, due to differences in syntactic constructions. Decisive criteria for support-verb constructions have been found in many languages, but they are not exactly the same. Butt (2010) also draws this conclusion on the basis of a review of typological and diachronic literature. In contrast, the fuzzy definition based on the semantic weight of the verb is language-independent.

But the price to be paid for language independence is that you cannot tell if an item satisfies the definition or not. Then, to which class does it belong? In practice, in order to avoid fuzzy membership of classes, and uncertain inclusions between classes, fuzzy features must be replaced by clear-cut, binary models of them.

Proponents of fuzzy features rarely claim these might be useful for computer applications. Semantic analysability/decomposability, for instance, is relevant to the mental lexicon, maybe to first language acquisition, but probably not to computational applications. If, in the future, information is attached to parts of idioms in formal semantic representations, can it be exploited computationally? Nunberg *et al.* (1994: 501) cite a variation on *tilt at windmills* ‘fight against imaginary or invincible opponents’:

(22) *Tilting At the Federal Windmill*

One can imagine that a parser that would handle a semantic representation of *tilt at windmills* composed of separate structures for the parts *tilt* and *windmill* might interpret (22) more correctly than a parser with an atomic semantic representation of the idiom. But, as of 2017, both types of parsers are hypothetical, since few parsers interpret idioms. The challenge of identifying even their least creative variants, such as *tilt bravely at windmills*, is probably a priority as compared to that of unlexicalized, playful variations like *tilt at a federal windmill*. Besides this remote perspective, computational applications of analysability are elusive. Bond *et al.* (2015) do not say that the analysability encoded in the dictionary is used by the English HPSG grammar to check restrictions to the form of the idioms.

[15] The use of *light verb* is loosely correlated with the intuitive approach and *support verb* with the technical approach. Gross (1981: 12) adopts the term of *support verb*, with the idea that, for example, *make* “supports” a predicational noun in (13a).

No clear use has been found either for another fuzzy feature, which consists in the fact that native speakers that don't know the MWE can guess its sense or not when they hear it in an uninformative context. For example, according to Osherson & Fellbaum (2010: 3), the sense of *rest on one's laurels* 'keep from making effort out of self-satisfaction from prior achievements' is easily guessed, but that of *walk on eggshells* 'be very cautious' is not.[16] This feature is relatively fuzzy: "the classification suggested above is only approximate, a number of idioms (...) cannot be straightforwardly fit into any one category"; they do not claim any computational use of this feature, and it is hard to imagine a realistic one.

In contrast, many clear-cut features are straightforwardly exploitable in language processing, especially those that directly determine the possibility of occurrence of actual forms, such as selected prepositions or applicability of syntactic operations. Clear-cut features of MWEs, as described in LGs, have been used early and recently in tree-adjoining grammars (Abeillé 1988), symbolic machine translation (Danlos 1992), finite-state parsers (Senellart 1998), symbolic dependency parsers (Tolone 2011) and statistical parsers (Constant *et al*. 2013).

Clear-cut features have significant methodological and practical advantages. When fuzzy features are used instead, it is worth checking carefully that their linguistic relevance motivates this choice.

# 3 Correlated features

## 3.1 An example

In some cases, it is easy to replace a single fuzzy feature with a bundle of clear-cut ones. For example, the syntactic operations applicable to verbal idioms, instead of being collectively considered as a single (fuzzy) feature, are better dealt with separately from one another.

Here are a few examples of syntactic operations.

• Optionality of fixed constituents:

(23) a. *John bearded the lion in his den*
'John faced the danger directly'

b. *John bearded the lion*

• Optionality of free slots:

[16] This feature is the more restrictive of the two features that Nunberg *et al.* (1994: 495) call *conventionality*. Osherson & Fellbaum term it *non-compositionality*, but it is quite different from *lack of compositionality* (cf. footnote 13) in Danlos & Gross (1988: 128-129): when speakers that don't know *rest on one's laurels* figure out what it means, they don't base their guesses only on a knowledge of the rules that determine the meaning and use of the parts when they occur separately, but also on their imagination and cultural familiarity with classical antiquity. Nunberg *et al.* (1994: 496–497)'s *transparency* (speakers' ability to guess why an expression with some literal meaning is used to convey a given non-literal meaning) is less restrictive: even if speakers can't figure out the non-literal meaning of an unknown idiom, they may be able to guess the motivation of this meaning once they are informed of it.

(24) a. *John bears comparison to Magritte*
‘John is similar enough to Magritte to be likened to him’

b. *John bears comparison*

• Insertion of free adjuncts:

(25) a. *This dealt a blow to my hopes*

b. *This dealt a strong blow to my hopes*

• Topicalization:

(26) a. *He would not deal such a blow to yo*

b. *Such a blow, he would not deal to you*

• Dative shift:

(27) a. *This dealt a blow to my hopes*

b. *This dealt my hopes a blow*

• Reduction in a repeated occurrence:

(28) *I changed my mind about China. You changed yours about India*

• Pseudocleft construction:

(29) a. *John fights fire with fire*
‘John uses the same arms as his opponents’

b. *The way John fights fire is with fire*

• Passivization:

(30) a. *The price of the coffee caught John short of change*
‘Given the price of the coffee, John had no change’

b. *John was caught short of change by the price of the coffee*

Not all operations are applicable to the same idioms. For example, *bear comparison to* admits the removal of its prepositional free slot, but not passivization:

(31) **Comparison to Magritte is borne by John*

Fraser (1970: 34) is aware of these differences between features. The straightforward model for syntactic flexibility is a multidimensional space of variation, since there are independent features. Nunberg *et al.* (1994: 509) claim that some large range of syntactic operations is “loosely correlated”. Contrasts such as (24a) vs. (31) show that such correlation, if it exists, is not 100%. Since correlation is a statistical notion, only statistical evidence could support Nunberg *et al.* (1994)’s claim. This could be done by measuring the correlation with extensive lexical data. For example, Freckleton (1985)’s table C1P2, which contains *beard the lion in his den*, shows a positive but loose correlation between optionality of the prepositional object and passivizability: the Pearson’s correlation coefficient, computed with these data, is 0.13.

Sag *et al.* (2002: 6), Baldwin & Kim (2010: 278) cluster all syntactic operations into SYNTACTIC FLEXIBILITY. Syntactic flexibility is an imprecise feature, since idioms that undergo only some of the known syntactic operations are intermediate cases, “syntactically flexible to some degree”, as Sag *et al.* put it. But they do not measure the correlation either. Even so, both Sag *et al.* (2002: 3) and Baldwin & Kim (2010: 279, Figure 12.1) derive some of their major classes from “the” feature of syntactic flexibility. Baldwin & Kim subdivide verbal idioms into two subclasses: one of non-decomposable idioms “with hard restrictions on word order and composition”, i.e. no application of syntactic operations, and another of decomposable, syntactically flexible idioms. They leave open the question of where intermediate cases should belong in practice: when some syntactic operations apply and others don’t, they cross-classify, i.e. they assign the same lexical item to different classes. None of these authors shows the benefit for NLP or linguistics of using a unidimensional scale as a model for a multidimensional variation space.

These models artificially create fuzzy features and the associated problems. Their authors do not explain what motivates this innovation, nor do they state a position on previous classifications that avoid equating distinct syntactic operations.

## 3.2 Discussion

Syntactic flexibility is a cluster of loosely correlated features and should not be used as if it were a single feature, especially when defining major classes: such definitions are imprecise. If a lexical database registers *bear comparison to* in a class of syntactically flexible idioms, this signals that this idiom admits at least some syntactic operations, but users cannot be certain about any specific operation, for example the removal of its prepositional free slot: (24b) *John bears comparison*. Reversely, if it is in a class of non-syntactically-flexible idioms, it does not admit all the possible operations, but users cannot safely deduce that it does not admit, say, passivization: (31) **Comparison to Magritte is borne by John*. This compromises computational usage: a major function of a classification is to ensure that the members of each class have the corresponding defining features.

Thus, as long as all properties are not securely established for all entries, it is a good practice to specify each criterion accurately. This leads to individuating a number of features, and to specifying which entries have which features, like in LG tables, which show that syntactic operations are not visibly more correlated in French (Gross 1982), Italian (Vietri 2011) or Greek (Fotopoulou 1993) than in English (Freckleton 1985).

A correlation between features may give a sense that they are particularly relevant to classification, because they might stem from a hidden, underlying, fundamental property. Moreover, if these features are used jointly for classification, assigning an entry to a class will implicitly specify all features at the same time, killing two birds with one stone. However, the temptation should be resisted until a systematic investigation assesses how much correlated the features are. If intuition overestimates the degree of correlation, which often happens, and if a classification equates loosely correlated features, assigning an entry to a class does not specify any of the features, failing to kill any of the proverbial birds.

Fundamental scientific progress has often been achieved by elaborating a distinction between two notions that are easy to confuse, e.g. weight and mass in physics. In their model, Sag *et al.* (2002) and Baldwin & Kim (2010) do the reverse: they replace a set of relatively precise features, which are objectively distinct and had been treated as such before, with an imprecise one, which is therefore more difficult to handle. Their model adds artificial uncertainty.

Thus, merging a cluster of loosely correlated features into an “aggregate” feature decreases the accuracy of the model and weakens its information content.

# 4 Reproducibility of observation of features

## 4.1 Examples

Even without any excess of optimism about correlations, some features are more clear-cut than others for another reason: reproducibility. Reproducible observations are those inherently susceptible to high inter-judge agreement. This notion may sound technical and is often ignored, but my point in this part is that it has considerable practical significance for projects with realistic goals.

Features are not equal in terms of inter-judge agreement. For example, the compulsoriness of a coreference relation in a MWE, e.g. in *think on one's feet* 'improvise a reaction quickly', between the subject and the possessive, is judged by checking the grammaticality or acceptability [17] of some sentences, which is relatively factual, as in (16) cited here as (32):

(32) a. *He thought on his feet*
b. **He thought on Ann's feet*

With *melt someone's heart* 'make someone feel sympathy', such coreference is not compulsory, as shown in (17a) *He melted your heart.* Few native speakers will disagree with such observations. Consequently, this feature can be recorded in a lexical database in a relatively reliable way. Similarly, the applicability of syntactic operations to MWEs is tested by applying the operations, as in (16), (17) and (23)–(31), while judging the acceptability of the result and the conservation of the meaning. Therefore, in most cases, it is also reproducibly observable.

In contrast, semantic analysability/decomposability in the sense of Gibbs & Nayak (1989), Nunberg *et al*. (1994: 508), Sag *et al.* (2002) and Baldwin & Kim (2010) has no other empirical ground than pure semantic impressions. "There are no well-defined procedures for specifying whether a given idiom is semantically decomposable or not" (Gibbs & Nayak 1989: 106). Nunberg *et al.*'s intuitions about the analysability of *take advantage of* are unstable (cf. §2.2). They cite *take stock* 'take the time to think' as analysable and *take hold* 'grasp' as non-analysable (p. 524), but other speakers' introspection is not sure to reproduce this contrast. According to them, *take stock* "can be roughly paraphrased as 'make an assessment', with the noun *stock* semantically approximating 'assessment'." But in the same vein, *take hold* can be paraphrased as 'voluntarily acquire a grasp', where *take* would denote the 'voluntary acquiring' and *hold* the 'grasp'. Acceptability is judged by introspection too, but is more factual and can be backed by corpus attestations in some cases. Joint use of introspection and corpus attestation is more and more recognized as a valid source of empirical data on acceptability (Johansson 1991: 313;[18] Fillmore 1992: 58, 2001: 1;[19] McEnery & Wilson 1996: 16;[20] Kepser & Reis 2005;[21] Gries 2011: 87 [22]). *Roll one's eyes*, in

---

[17] By *grammatical* we mean that a sentence may be used to convey some information in some situation and in some context. This is consistent with how *grammatical* is used by most linguists, and identical to what Harris (1957: 293) means by *acceptable* (Ross 1979: 161). In fact, we will use *acceptable*, to avoid confusion with Chomsky's use of *grammatical* (Chomsky 1957: 15), which is in principle divergent since some nonsense sequences may be grammatical in his sense.

[18] "The corpus remains *one* of the linguist's tools, to be used together with introspection and elicitation techniques. Wise linguists, like experienced craftsmen, sharpen their tools and recognize their appropriate uses."

[19] "One [cannot] success in the language business without using both resources: any corpus offers riches that introspecting linguists will never come upon if left to their meditations; and at the same time, every native speaker has solid knowledge about facts of their language that no amount of corpus evidence, taken by itself, could support or contradict."

[20] "Why move from one extreme of only natural data to another of only artificial data? Both have known weaknesses. Why not use a combination of both, and rely on the strengths of each to the exclusion of their weaknesses? A corpus and an introspection-based approach to linguistics are not mutually exclusive. In a very real sense they can be gainfully viewed as being complementary."

[21] "It is one of the main aims of this volume to overcome the corpus data versus introspective data opposition and to argue for a view that values and employs different types of linguistic evidence each in their own right."

[22] "It is obvious that corpus linguists need to make subjective decisions all the time, and they need to document

its idiomatic meaning, denotes a ‘feeling of surprise and rejection for something stupid or strange’, and often also an actual eye movement that expresses this feeling, but this physical element of meaning is perhaps not necessarily present in all occurrences of the idiom:

(33) *We’ve all rolled our eyes at a particularly catchy headline*

Many idioms share this property (Burger 1998: 44). In their meaning, the feeling part is rather non-analysable, whereas the physical movement part is rather analysable. When compromising between these intuitions, not all speakers are likely to obtain the same result. It is not just that the semantic analysis of the feeling part is funny: more importantly, there is no reason why different observers would assign the idiom to the same class.

The contrast between more or less reproducibly observable features is also observed in French, in Italian, and presumably in any language. Take this French idiom:

(34) *se mettre le doigt dans l’œil*
(lit. put one’s finger in one’s eye)
‘have a mistaken understanding’

My own impressions in terms of analysability are precarious: does *le doigt* (lit. one’s finger) really stand for an element of meaning like ‘understanding($x$)’, *mettre* (lit. put) for ‘choose’ and *dans l’œil* (lit. in one’s eye) for ‘wrong’?

Semantic analysability poses recurrent problems of reproducibility of observation. This makes it a fuzzy feature.

## 4.2 Related work

Reproducibility of observation is not a new requirement. It is a central concern for American structuralists such as Bloomfield and Harris, who typically improve it by adjusting the definition of features under analysis, and in particular by resorting to FORMAL OR SYNTACTIC CRITERIA, as in (16b) **He thought on Ann’s feet*, avoiding to rely directly on pure semantic intuition. This tradition focuses on selecting knowledge that can be reproducibly observed, as part of a quest for scientificity in linguistics. In the observation of semantic features, DIFFERENTIAL SEMANTIC EVALUATION is more reproducible than absolute semantic evaluation (Gross 1975: 391–392).[23] For example, take the following French support-verb construction:

(35) *Le mur a de la couleur*
(lit. The wall has colour)
‘The wall is colourful’

Far from all interviewed speakers agree that (35) denotes intensiveness; in other words, this observation is little reproducible. Now take the following variant:

(36) *Le mur a une couleur*
‘The wall has a colour on it’

their subjective choices very clearly in their publications. However, in spite of these undoubtedly subjective decisions, many advantages over armchair linguistics remain: the data points that are coded are not made-up, their frequency distributions are based on natural data, and these data points force us to include inconvenient or highly unlikely examples that armchair linguists may ‘overlook’.”

[23] “Pairs of sentences that are candidates for being related by a transformation are judged to be synonymous or not. Thus, meaning is only involved in comparisons, and *differences* in meaning are detected in this manner. In the physical sciences, it is well-known that *absolute* evaluations of a variable (e.g. temperature) lead always to rather crude results, when compared to *differential* evaluations of the same variable. The situation appears to be the same in linguistics with respect to meaning. Attributing absolute terms to forms is quite problematic, and anyway, has proved to be rather unsuccessful, while comparing the meanings of similar forms may bring to light subtle differences that may be hard to detect directly.”

When asked if (35) is more intensive than (36), much more speakers share this perception, agreeing that (36) is more neutral. The differential observation is more reproducible than the absolute one.[24] Reproducibility decreases back if you compare phrases, for example *de la couleur* vs. *une couleur,* instead of complete sentences. The LG method is much about such practical techniques of elaborating the procedures of observation or the definition of features, in order to improve reproducibility. When you ask the right question, it is easier to agree on an answer. In practice, performers of LG work are trained to be systematically watchful of their own dubious or instable judgments, and to compare these judgments to those of their peers. This measurement of reproducibility is subjective, but peer controlled, in order that subjectivity does not affect the quality of the results. It is performed right from the beginning of the project, and separately for each feature, to detect which features raise reproducibility problems. Such detection leads to two types of decisions:

(i) give up the study of 'bad' features, that is those that cannot be observed with reasonable reproducibility

(ii) look for 'good' features for which methodological precautions ensure reasonable reproducibility

A reproducibility issue often causes a shift from an intuition-defined feature to one or several new criterion-defined ones. The latter may be a little different, but they have an advantage: it is clearer what they are. Such decisions refine or shift the target of the description and, of course, eventually affect the classifications based on the features.

There are few ongoing debates about such practices in linguistics, and even less in research on MWEs. Reproducibility is alien to Baldwin & Kim (2010)'s concerns, except for an allusion in connection with interpretation of nominal compounds (p. 275).[25] In current practices beyond LG, measurement of reproducibility is more objective: it takes the form of inter-judge agreement statistics. But these statistics either focus on a small sample of features deemed representative, or handle features collectively, not individually (Palmer *et al.* 2005: 86 is an exception): in both cases, they don't help to tell the 'good' features from the 'bad'. The inter-judge agreement approach tests how a team of descriptive linguists fare as regard reproducibility, it does not assess the potential of each feature. It views reproducibility as a behavioral problem only, not as a syntactic or lexicological problem, and disregards the fact that the problem is different for each feature.

In addition, current practices usually take into account only small samples of prototypical MWEs deemed representative.[26] But many reproducibility issues stem from the diversity of lexical entries: their detection requires comprehensive scrutiny of the lexicon.[27]

Globally, with the shift from subjective to objective procedures, quality of measurement has deteriorated. It is worse than that: now, reproducibility assessment is rarely used for feedback on the aims of the description or on its practical procedures. First, such feedback would require differential assessment on individual features. Second, inter-judge agreement is usually computed in the end of the descriptive phase (Meyers *et al.* 2004: 803 is an exception), when it is too late for feedback. Reproducibility assessment is only regarded as a quality indicator: researchers are content with measuring the symptoms and seldomly attempt to cure them.[28]

[24] Here are two other examples of definite semantic differences: (35) denotes more of a favourable subjective judgment than (36), and (35) may evoke one or several colours, whereas (36) evokes one.

[25] Anyway, in the case of analysability, no such improvements of the definition seem to be at hand.

[26] Gibbs *et al.* (1989: 60) assess the consistency of undergraduates' judgments of semantic decomposability of a sample of 36 idioms.

[27] The complexity of the assessment of reproducibility has three dimensions: the number of features, the number of lexical entries and the number of judges. Informal LG practices deal with all three dimensions. But objective measurement of inter-judge agreement is costly, which leads to limiting its ambition in terms of two of the three dimensions: the number of features and of lexical entries. Thus, the operation loses its essential benefits.

LG also contributes to reproducibility indirectly, by supporting the publication of results in readable formats. LG tables display readably which entries have which features. Their well-known tabular format is theory-, framework-, formalism- and implementation-independent and allows for explicit negative information, e.g. the fact that *bear comparison to* has no passive. Publishing LG tables in scientific publications and web sites indirectly tends to increase reproducibility, since peers can easily check if they agree with the recorded information. Kaalep & Muischnek (2008) adopt a readable tabular format too, but do not individuate columns for individual features. LG tables are used as source code, i.e. for manual edition; they usually need to be automatically translated into application-dependent formats (Tolone & Sagot 2009; Constant & Tolone 2010), and this is their main flaw for computational linguists (Hathout & Namer 1998; Gardent *et al.* 2005).[29] However, other formats are less readable: the DuELME (Grégoire, 2010: 34–36) and Lefff (Tolone & Sagot 2009) formats contain lists of features without explicit negative information: to check that an entry does *not* have a given feature *f*, you have to verify that none of the features it has is *f*.

## 4.3 Discussion

Reproducibility is an epistemological requirement for scientificity. Low reproducibility casts a doubt on what exactly a feature is, since different observers perceive the feature differently. Gross (1981: 14) even says about traditional semantic classification of prepositional adjuncts: “Perception of such distinctions depends a lot on individuals; therefore, they might be of no interest” (my translation). Features with high reproducibility of observation are essential when lexical entries are described manually, and provide a good basis for a classification with an ambition of stability and scientificity. In addition, many of these features are factual, and therefore informative for further research, no matter the linguistic theory adopted. Factual features are valuable for language processing too, especially when they determine the possibility of occurrence of actual forms, as in (16), (17) and (23)–(31): this is essential to automatically recognizing such MWEs.

LG authors’ experience on a number of languages proves that the requirement of reproducibility does not drastically limit the diversity of features to be studied. Their tables of MWEs contain a large collection of useful features. For instance, the study of prepositional-phrase idioms compatible with *be* in English by Machonis (1987) showed that many of them admit a syntactic operation that inserts verbs such as *get* or *throw* and a causative or agentive subject, as *be in a jam* ‘be in trouble’:

(37) a. *Kathy was in a jam*
b. *An unfortunate situation had* (*got* + *thrown*) *Kathy into a jam* [30]

---

[28] Another way of improving the resources has emerged: automating error detection targeted at specific error types in resources. For example, Meyers *et al.* (2004) automatically check formal and heuristic properties of dictionary entries; Cohen *et al.* (2011) check a constraint that the occurrences of dictionary entries in a corpus are supposed to fulfill. These a posteriori checks are celebrated as contributing to “quality assurance” and to “the development of a true science of annotation” (Cohen *et al.* 2011: 82), but they are hardly relevant to the present discussion, since they do not target features with reproducibility issues. In addition, they do not contribute to refining the target of description, as a priori vigilance about reproducibility does.

[29] Tolone (2011) improved part of the LG in that regard: she homogenized the mnemonic identifiers of properties, encoded properties common to whole classes and created a user documentation in English.

[30] The notation (*got* + *thrown*) serves to refer to several variants, here both *had got Kathy into a jam* and *had*

But some idioms don't admit this operation with the same verbs, as *be in the wrong* 'be morally or legally wrong':

(38) a. *The cyclist is in the wrong*
b. *Slapping the pedestrian* (*got* + **threw*) *the cyclist in the wrong*

Features related to this little-known causative construction, usually classified in recent literature under the large category of "lexical variation", are decisive for the automatic parsing of sentences such as (37b).

## 4.4 A more detailed example

Semantic features such as analysability are rare in LG descriptions: they are difficult to define with sufficient rigour. I will exemplify this difficulty with a new semantic feature which is interesting for NLP, but requires a precise definition before it can be encoded in LG tables.

The French law term *citer un témoin* 'call somebody as a witness' (lit. quote a witness) is a MWE because the verb *citer* has this meaning only with the noun *témoin*. Still, the meaning of this noun in the idiom is the same as a meaning this noun can also have, as a (lexicalized) law term, when *citer* is not present at all in the context. Even in the idiom, it usually refers to a specific person. It can belong to a chain of coreferring expressions, no matter whether it is the first element of the chain, as in (39), or not, as in (40).

(39) *La défense a cité un témoin. Il vient de s'exprimer*
(lit. The defence quoted a witness. He has just expressed himself)
'The defence called a witness. He has just spoken'

In (39), *un témoin* 'a witness' and *il* 'he' refer to the same person.

(40) *Ils avaient un autre témoin, mais finalement ils ne l'ont pas cité*
(lit. They had another witness, but finally they did not quote him)
'They had another witness, but they ended up not calling him'

In (40), *un témoin* in the idiom is replaced by the pronoun *l'* 'him', which refers to the same person as *un autre témoin* 'another withess'. In a chain of coreferring expressions like those of (39) and (40), the syntactic markers of the coreference such as determiners, pronouns, etc., follow the same rules as when the noun is not part of an idiom. For example, in (39), *il* 'he' has the same form as when *un témoin* 'a witness', but not the rest of the idiom, is present in the context:

(41) *La défense a un témoin. Il vient de s'exprimer.*
'The defence has a witness. He has just spoken'

The feature that I wish to single out, and which *témoin* 'witness' in (39) shares with many other idiom components, is a combination of three properties:

(i) The component, when used in the idiom, has mandatorily a meaning it can also have (as a lexicalized meaning) when the rest of the idiom is not present at all, not even in the context, as opposed to *feet* in *think on one's feet* 'improvise a reaction quickly', or to *eyes* in *roll one's eyes*, where *eyes* doesn't always refer to eyes.[31]

---

*thrown Kathy into a jam*. This notation inspired from algebra and commonly used in LG is more informative than the notation *got/thrown*, since the parentheses delimit precisely where each variant begins and ends.

[31] Property (i) matches what Burger (2007: 96) calls *partly idiomatic* expressions. It is more restrictive than analysability/decomposability: for instance, in *pull strings* 'covertly use one's influence on personal connections', the noun *string* does not keep any of the lexicalized meanings it has when *pull* is not present at all. As a consequence, the feature that I am defining is different from analysability too.

(ii) The component can be the first in a chain of coreferring expressions, and then the syntactic markers of the coreference: determiners, pronouns, etc., follow the same rules as when the noun is not part of the idiom. This does not happen, for instance, with *posture* in the French idiom of (42):

(42) *Kathy était en mauvaise posture*
(lit. Kathy was in bad posture)
'Kathy was in trouble'

To refer to the trouble after they have used this idiom, speakers use another noun:

(43) *Kathy était en mauvaise posture. Ces difficultés auraient pu être évitées*
(lit. Kathy was in bad posture. This trouble could have been avoided)
'Kathy was in trouble. This trouble could have been avoided'

Without this idiom, they can use the same noun:

(44) *Kathy avait une posture fière. Cette posture a été commentée*
'Kathy had a proud posture. This posture has been commented'

But if the first expression referring to the trouble is part of the idiom of (42), speakers do not use the same noun for other coreferring expressions:

(45) **Kathy était en mauvaise posture. Cette posture aurait pu être évitée*
(lit. *Kathy was in bad posture. This posture could have been avoided)
'Kathy was in trouble. This trouble could have been avoided'[32]

(iii) The component can occur in a chain of coreferring expressions without being the first, and then the syntactic markers of the coreference such as determiners, pronouns, etc., follow the same rules as when the noun is not part of the idiom. This does not happen, for example, with *strings* in *pull strings* 'covertly use one's influence on personal connections'. When speakers refer to the connections before using this idiom, the coreference between the first mention of the connections and the idiom component is not explicitly marked:

(46) *I needed connections to make myself known, and John could pull strings for me*

In this form, *strings* has the same form as if there were no mention of it before. Without the idiom, we observe a syntactic marker of coreference:

(47) *I needed connections to make myself known, and John provided them to me*

Speakers do not use this marker if the second mention of the connections is part of the idiom:

(48) **I needed strings to make myself known, and John could pull them for me*

But why get interested in the combination of features (i)-(iii),[33] which has never been studied or named? Because it is shared by many other terminological idioms, for example the French term of geometry *abaisser une perpendiculaire à* 'drop a perpendicular on to' (lit. move down a perpendicular to).[34] The idiom component that has the feature, like *témoin* 'witness', is often

---

[32] Example (45) is not entirely parallel to (39): (39) involves pronouns and (45) involves determiners and nouns. Studying feature (ii) requires taking into account diverse syntactic markers of coreference. This feature is connected with pronominalizability, but not only.

[33] It is not an "aggregate" of features (i)-(iii) as in §3 above: it is specifically a conjunction of these independent features, in the sense that an idiom has it if and only if it has simultaneously (i), (ii) and (iii). Alternatively, features (i)-(iii) might be studied separately, but they are likely to be less useful than their conjunction.

[34] *Abaisser* has this meaning only with *perpendiculaire* and *parallèle* 'parallel', which keep their autonomous terminological meaning and ability to be referred to anaphorically:

(i) *On abaisse une perpendiculaire de A à BC. Cette droite est parallèle à CD*
'A perpendicular is dropped from A on to BC. This line is parallel to CD'

a technical term too, and is able to denote a referent in a clear and specific way. In such case, these idioms are meaningful elements of technical texts, a realistic target for future improvements to the automated understanding of natural language texts.

My definition is based on properties (i)-(iii), which are relatively formal.[35] Even so, this feature is probably not ready for encoding, that is for production of a satisfactory list of the idioms with this feature: only large-coverage encoding experiments would tell if this definition ensures sufficient reproducibility of observation.

In this section, all the examples above focus on nouns that are parts of verbal idioms. How does the feature extend to other PoS? Here are two noteworthy features closely related to this one.

In a large proportion of multiword nouns, the head noun keeps all the grammatical and semantic behaviour it has as an independently existing lexical entry. This is the case of *red wine*: it is a (terminological) MWE because *red* has this meaning only with *wine*, but *wine* can be equated with the independent lexical entry *wine,* with the same properties (i)-(iii) as above. With multiword nouns, this is related to another test: *red* can usually be inserted in sentences with *wine* or removed from them, without unexpected changes in acceptability or meaning:

(49) *They have an interest in wine*
*They have an interest in red wine*

(50) *Is red wine healthy and worth the calories?*
*Is wine healthy and worth the calories?*

This test uses differential semantic assessment. *Smooth operator* 'persuasive person; manipulative person' doesn't share this feature, as the meaning changes in (51)–(52) show:

(51) a. *Ask the operator to dial*
'Ask the switchboard operator to dial'

b. *Ask the smooth operator to dial*
'Ask the persuasive person to dial'; 'Ask the manipulative person to dial'

(52) a. *Any lady I've dated will tell you I'm no smooth operator*
'Any lady I've dated will tell you I'm not manipulative'

b. *Any lady I've dated will tell you I'm no operator*
'Any lady I've dated will tell you I'm not a switchboard operator'

Some adverbs are specific to one or a few verbs, which nevertheless keep all their behaviour. Here are two examples in French from the tables of multiword adverbs by Gross (1986):

(53) a. *(chanter + crier + rire) à gorge déployée*
(lit. (sing + shout + laugh) at opened-out throat)
'(sing + shout + laugh) out loud'

b. $N_0$ *aller à* $N_1$ *comme un tablier à une vache*
(lit. $N_0$ fit $N_1$ as an apron does a cow)
'$N_0$ fit $N_1$ supremely badly'

[35] My definition does not include another striking property of *témoin* 'witness' in (39): it can refer to a specific entity, as opposed to *fire* in (29a) *John fights fire with fire*, which alludes to ways of fighting in general. The semantic distinction between specific and generic reference is a matter of pure intuition. As such, its reproducibility of observation can be low, for example in the case of *bread* in *They will take the bread from our mouths* 'They will divert money from us': in this sentence, does *the bread* refer to material goods in general, or to a specific instance of income?

As opposed to the preceding examples, these have no terminological value.

Care for reproducibility in observation of linguistic facts characterizes a conception of humanities in which scholars not only share insights and deepen their intuition, but also gather reliable factual knowledge, paying attention to practical techniques that improve the quality of their description. In this conception, descriptive work, and in particular lexical description, is fundamental. For instance, assessing reproducibility of observation is a practical matter: it involves scanning through the lexicon while trying to describe which entries have a feature and which don't. Thus, preferring features that can be observed by humans in a reproducible way is good practice.

# 5 Checking information against the lexicon

## 5.1 Discussion

Checking information against the lexicon is still alien to a large part of MWE research. Baldwin & Kim (2010) do not cite studies using intensive lexical description of MWEs, except for Estonian MWEs. The companion website to their paper [36] cites corpora and tools, but no NLP dictionaries; among the tools, it even omits those based on dictionaries. Current descriptive research is not eager to achieve large lexical coverages. FrameNet has a small coverage of MWEs (Hartmann & Gurevych 2013), and so do, among NLP dictionaries, VerbNet, WordNet and Meaning-Text. Bond *et al.* (2015) encode for HPSG a sample of English idioms with a possessive coreferent with the subject, like *roll one's eyes*, but here is how the size of the result compares with previous efforts: Freckleton (1985)'s classes C1A and C11 contain 538 verbal idioms with such a possessive, while Bond *et al.* (2015: 64)'s four classes that correspond to C1A and C11 total 168 ones. Aside from LGs, sizable lexical databases of MWEs are few. The NomLex-Plus and NomBank dictionaries of English nouns with predicate-argument structure list 8000 entries (Meyers, 2007). Kaalep & Muischnek (2008)'s database lists 13000 Estonian MWEs. The DuELME dictionary of Dutch MWEs totals 5000 expressions (Grégoire, 2010).[37]

My point in this part is that information on MWEs is worth checking against the lexicon. Reluctance against lexical description is rarely explicit, and when it is, it is not motivated by sound reasons.[38]

Sure, intensive investigation into the lexicon is costly. For example, the construction of LG tables of MWEs, which are comprehensive repositories with representation of individual features (cf. §2.2.1), has always involved considerable work. But the objective of a satisfactory processing of MWEs is worth cost and effort. (The reason one enjoys Dostoevsky is not because he is easy to read.) And the tables are available for several languages, which shows that this work is realistic.

---

[36] http://handbookofnlp.cse.unsw.edu.au/?n=Chapter12.Chapter12 looked up in August 2016.

[37] The SemLex Dictionary of Czech MWEs is still little documented in publications (Bejček & Straňák 2010).

[38] "Again, in itself this type of approach [interviews, surveys, statistics] is neither good nor bad. The question is whether it leads to the discovery of principles that are significant. We are back to the difference between natural history and natural science. In natural history, whatever you do is fine. If you like to collect stones, you can classify them according to their color, their shape, and so forth. Everything is of equal value, because you are not looking for principles. You are amusing yourself, and nobody can object to that. But in the natural sciences, it is altogether different. There the search is for the discovery of intelligible structure and for explanatory principles. In the natural sciences, the facts have no interest in themselves, but only to the degree to which they have bearing on explanatory principles or on hidden structures that have some intellectual interest" (Chomsky 1979: 58-59). Beyond the depreciative rhetoric of this passage, Chomsky actually suggests skipping factual observation when it involves extensive description.

The reluctance towards intensive lexical description might come from a feeling that it is deemed an unskilled, low-grade occupation. But such a feeling is unfounded: in projects of construction of large lexical databases of MWEs, linguists are obviously engaged in highly skilled labour.

The reluctance may be directed towards manual work. As computer science is about automating information processing, many computational linguists may understandably feel excited about devising "knowledge-free" solutions that avoid the need of *any* labour-intensive activity, be it in preliminary operations. But, in the case of MWE-related NLP, relying on this only hope is adventurous: the goal of fully automating acquisition of knowledge about all MWEs has been giving hard times to the community during more than 15 years.

No dictionary is 100% complete or 100% error-free, but this does not make them useless.

And manual lexical description has several advantages. The resulting data allows for more well-documented studies and is likely to be useful for making successful rules or devising successful machine learning experiments. When linguists scrutinize 10 features on a comprehensive part of a 1000-item class, what they find out is worth taking a look. It provides examples and counter-examples which are useful to test predictions, proposals and hypothetical rules or generalities. LG tables, as large repositories of factual features, are a source of examples for further research, no matter the theory, framework or implementation to be used. Creativity of language is a major obstacle to its scientific study, and it lies, among other things, in the combinatorics of lexical items and grammatical constructions: systematic investigation in the lexicon is therefore a way of addressing this problem.

Intensive lexical description is crucial to selecting features for classification, and therefore to the quality of classification. The construction of the NomLex-Plus and NomBank dictionaries of English nouns with predicate-argument structure involved an unprecedented investigation into support-verb constructions in English and into features to recognize and classify them (Meyers, 2007). The study of idioms that take the form of a prepositional phrase in Romance languages (Danlos 1980; Ranchhod 1990; Gross 1996; Vietri 1996), English (Machonis 1988) and Greek (Moustaki 1995) singled out a particularly useful feature for the top of MWE classifications. Some of these idioms are compatible with *be* or the equivalent copula in other languages and may appear in predicative position, as in (54b):

(54) a. *John will reach the end on time*
b. *John will be on time*

Others may not:

(55) a. *The crisis has a demographic cause in the final analysis*
'The crisis has a demographic cause, when everything has been considered'

b. **The cause is in the final analysis*

Those compatible with *be*, like *on time* 'punctually; punctual' in (54), *on vacation* and *on the spot* 'immediately; in the same place; in trouble', usually pose a problem of PoS: are they closer to adjectives or to adverbs?[39] In contrast, those like *in the final analysis* in (55a) and *for instance* are clearly adverbial expressions. Compatibility with *be*, that is the contrast between (54) and (55), provides a relatively sharp division in a large number of cases where PoS distinction would otherwise be particularly uncertain. Applying this criterion requires investigating into the syntactic contexts of idioms in sentences, but this is the usual price to be paid to resolve PoS issues,[40] and PoS are key to a general classification of MWEs. So, this

[39] Lexicalized MWEs are lexical items, so they may have a PoS like single-word lexical items do.
[40] The most appropriate definition of each PoS is based on its possible syntactic contexts in sentences. For

criterion is more relevant than the presence vs. absence of a determiner, retained by Baldwin & Kim (2010: 278) at the top of their classification, a criterion that only uses the internal structure of idioms.

Lexical data deepens knowledge of how correlated two features are. It does so by providing reliable statistics on lexical entries: how many entries with feature *f* also have feature *g*? For example, the causative construction of (37b), with a prepositional-phrase idiom and *get, throw* or other verbs like *keep*, is observed only when the idiom is compatible with *be*:

(56) *John will be on time*
*This gift will keep John on time*

(57) **The cause is in the final analysis*
**This point keeps the cause in the final analysis*

Such specific grammatical information allows for measuring correlations accurately.

Intensive lexical description tends to make researchers more cognizant of variation, including less frequent variations and variations of less frequent items. As such, it is complementary to corpus annotation, which rather makes them aware of context-related issues.

Lexical description also provides means of separating homonymous entries, for example the various interpretations of *on the spot*: 'immediately'; 'in the same place'; 'in trouble'. Such separation, in turn, is essential to construct cross-lingual tables (Ranchhod & De Gioia 1996).

All these benefits of lexical description make it *a priori* useful for applications. There is still little significative feedback from the NLP use of any comprehensive dictionary of MWEs, but this may come from the complexity of the problem and the interdependence of all subproblems of symbolic syntactic parsing.

## 5.2 Predicted vs. checked features

Gibbs & Nayak (1989: 104) hypothesize that semantic analysability/decomposability "determines the syntactic behavior of idioms". In this section, I examine the present and potential consequences of this conjecture.

With Nunberg *et al.* (1994), the hypothesis becomes two claims. First, the analysability of an expression predicts syntactic operations are applicable to it: "the syntactic properties of idioms [that is the applicability of syntactic operations] are largely predictable from the semantically based analysis of idioms we are proposing [i.e. their analysability]" (p. 507). In parallel, the unanalysability of an expression predicts syntactic operations are not applicable to it: "we (...) explain a variety of 'transformational deficiencies' of idioms by positing a bifurcation between [unanalysable] and [analysable] expressions, with only the latter type permitting those processes" (p. 508).

After these claims, analysability became popular in the community and was used to define some of the major classes of MWEs. The two predictions give a sense that analysability is an underlying, fundamental property, and that its use in classification implements a strategy of parsimony, since assigning an entry to a class automatically specifies all the predicted features. Sag *et al.* (2002: 4) retain only the second prediction: "due to their opaque semantics, non-decomposable idioms are not subject to syntactic variability, e.g. in the form of internal modification (*#kick the great bucket in the sky*) or passivization (**the breeze was* example, in English, a noun is to be recognized by its ability to be preceded by determiners and adjuncts, followed by adjuncts, etc.

*shot*).”[41]

However, Nunberg *et al.* (1994) do not check the claims, either on available data or on original data. A general claim requires systematic verifications, which they mention as a perspective for future work: “testing this prediction systematically is a nontrivial project” (p. 531). Therefore, both claims remain hypotheses.

When authors check the predicted syntactic features, they readily find out counter-examples to both predictions (Abeillé 1995; Stathi 2007). Here are three more that I picked from the lists of French verbal idioms by Gross (1982):

(58) *rater un éléphant dans un couloir*
(lit. miss an elephant in a corridor)
‘be unable to hit the broad side of a barn, have poor aim, be unable to reach targets’

Example (58) seems analysable as *miss*(*x*, *easy-target*), but does not admit syntactic variations, not even omission of the prepositional complement.

(59) *trouver chaussure à son pied*
(lit. find shoe to one’s foot)
‘find the perfect match for oneself’

Example (59) seems analysable as something like *find*(*x*, *partner*), but does not admit syntactic variations either. Conversely, (60) is hardly semantically analysable:

(60) *mettre toutes les chances de son côté*
(lit. put all the chances on one’s side)
‘not take any chances’

But it admits the passive form:

(61) *Toutes les chances sont mises de votre côté*
(lit. All the chances are put on your side)
‘You are not taking any chances’

Nunberg *et al.* (1994: 512) extend their claims in the case when an idiom is analysable: “the syntactic versatility of an idiom is a function of how the meanings of its parts are related to one another and to their literal meanings”. In other words, details of the semantic structure of analysable idioms would predict which syntactic operations are applicable.

It is particularly difficult to give credit to this hypothesis. Its authors do not check it any more than the previous one; the alleged rules of prediction are unknown. Formalizing them would be a challenge that no one has taken up since. Instead, Riehemann (2001) finds that which types of syntactic variation a given idiom can undergo is highly unpredictable.

Baldwin & Kim (2010: 280) adopt Nunberg *et al.*’s hypothesis as their own: “the exact form of syntactic variation [of verbal idioms] is predicted by the nature of their semantic decomposability”. But they do not provide any evidence to support it. Even worse, their formulation suggests that, instead of describing the syntactic variation of verbal idioms, one might infer it automatically from a description of their analysability. But recall that syntactic variation is more reproducibly observable than analysability: thus, the suggested proposal comes down to inferring several factual features from a property that poses problems of definition and observation (cf. §3.2). Such a process would hardly be effectual.

Predicting features might seem a clever move. But it necessarily begins as a hypothesis, which needs to be checked to get any scientific value. So, predicting features does not allow

[41] *Shoot the breeze* means ‘talk casually’.

for bypassing the verification step.

# 6 An adapted classification

I propose in Figure 1 a decision tree adapted from Baldwin & Kim (2010: 279), but which avoids the flaws discussed above, and in particular features that are too fuzzy or difficult to observe. It uses all the MWE-related LG work, including the studies on English, Romance languages, Greek, Korean and other languages, cited in §2.2.1, §2.2.3, §3.2 and §5.1. Much of this work was conducted in parallel and cross-linguistic comparisons showed that, even though the details of formal criteria depend on the languages (§2.3), the notions they define are similar. For example, the typology of French support verbs by Gross (1998) is transferred to the English FrameNet by Ruppenhofer *et al.* (2006: 37–38) without any modification. The classification in Figure 1 is in terms of notions defined by criteria mentioned in the text of this chapter, but the criteria are not repeated in the figure. Thus, it is formulated for English and easily adaptable to many other languages: to adapt it to French, substitute *être* for *be*.

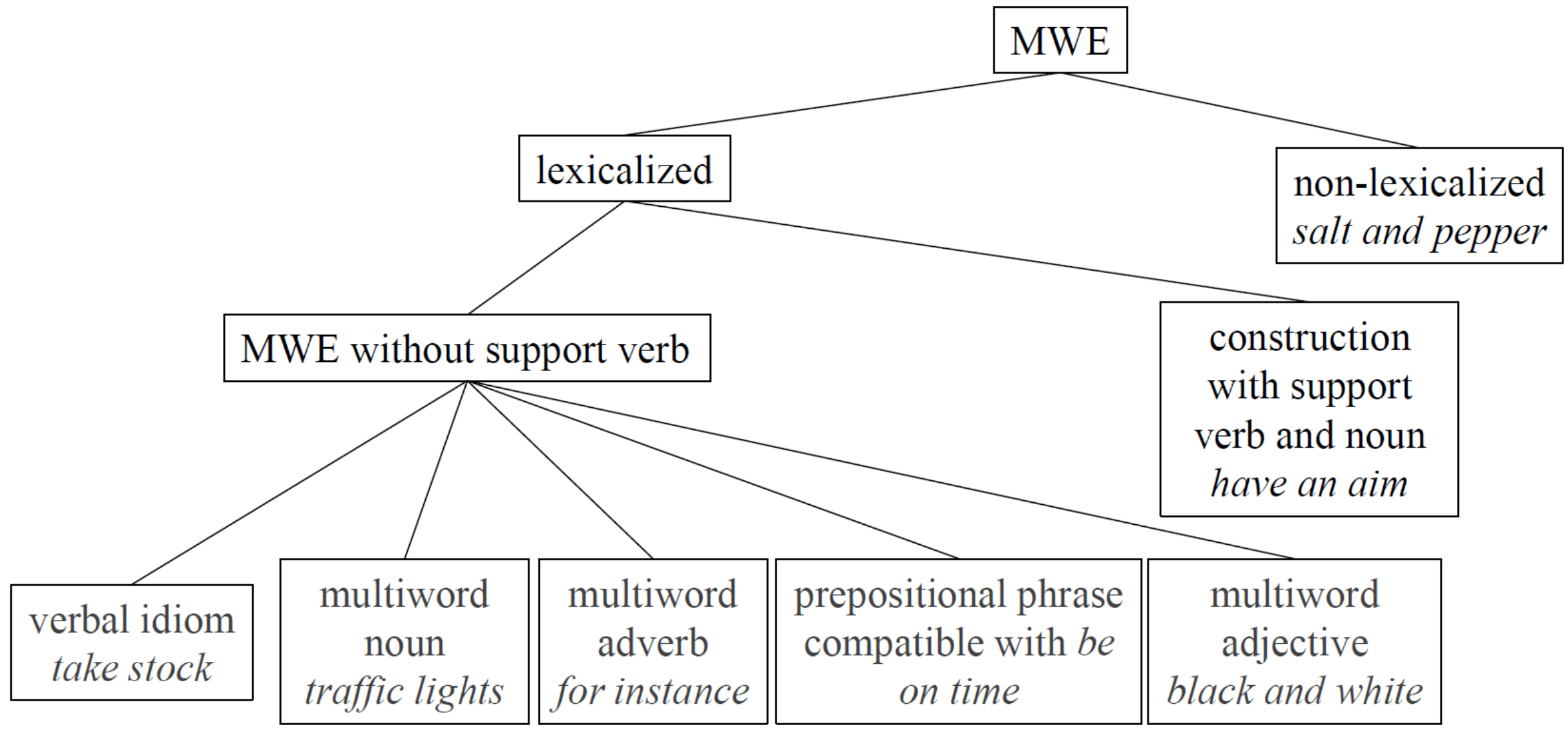

*Figure 1: Classification of MWEs.*

The top distinction is between lexicalized and non-lexicalized expressions. By non-lexicalized expressions, I mean those that are fully compositional but in which a statistical preference for an element is not explained by extra-linguistic facts. For example, the preference for *sell a house* over *sell a wall* is explained by cultural habits, so we don't need to describe it as a linguistic property; therefore, *sell a house* is not a MWE. In contrast, the preference for the French phrases *tondre la pelouse* 'mow the lawn' and *couper l'herbe* 'cut the grass' over *tondre l'herbe* 'mow the grass' and *couper la pelouse* 'cut the lawn' is a purely linguistic fact. This suggests *tondre la pelouse* and *couper l'herbe* are MWEs, but they are not lexicalized, since the other two are in use. The term *black and white* 'composed of shades of black or of a single colour' is lexicalized. If it were fully compositional, speakers would be able to interpret *white and black* the same way as *black and white*, which they aren't. The same holds for *traffic lights*: if it were fully compositional, speakers would be able to interpret it as another type of light connected with traffic.

The second distinction in Figure 1 relies on the notion of support-verb construction. This is not an easy distinction, but the literature shows that trained linguists are able to make it on the basis of formal criteria that ensure sufficient reproducibility of observation: these criteria are outlined in §2.2.3. Support-verb constructions are a significant class because they are numerous both in texts and in a dictionary: out of the 62,100 MWE entries of the French LG, 12,700 (20%) are support-verb constructions (Tolone 2011). Support-verb constructions have in common a crucial property which is a good reason to place them so close to the top of the classification: the construction with the verb, for example *have a passion*, and the construction without the verb, which is usually a predicational noun, here *passion*, are not adequately described by two distinct lexical entries. For example, the arguments of *have a passion* are exactly the same as those of *passion*, including the preposition of the complement (*for*) and the restrictions on what may fill both slots (the subject contains a human noun; the complement may contain a human, concrete or abstract noun or an infinitival clause). Moreover, occurrences without the support verb are usually more frequent in texts than occurrences with it (Laporte *et al.*, 2008).

The third distinction in Figure 1 is based on PoS.[42] It conflates adverbs with prepositions and conjunctions, with the view that a multiword preposition like *in spite of* or a multiword conjunction like *in case that* may also be analysed as a multiword adverb with a free prepositional or clausal slot.[43] This lowest level of the tree includes an additional "PoS", namely "prepositional phrase compatible with *be*", for instance *on time* and *on vacation*, in order to sort out partially the problem of assigning a PoS to idioms taking the form of prepositional phrases: are they adverbs or adjectives? The compatibility with *be* provides a relatively sharp division in a large number of cases where PoS distinction is particularly uncertain (cf. (54)–(55), §5.1). The multiword adjective category is meant for expressions that do not take the form of prepositional phrases, for example *black and white* 'composed of shades of black or of a single colour' or *safe and sound* 'unharmed'. In Figure 1, the distinction between support-verb constructions or not is just above the decision about PoS. It could also be the other way round, which would make the support-verb-construction class a brother of the verbal-idiom class. This variant would give prominence to PoS, which are always key information, well-known classes and often clear-cut features.

In Figure 2, I propose an alternative classification that may bewilder many researchers. But those that take seriously the notion of support-verb construction will probably find it more consistent than Figure 1. From Gross (1981: 34), Ranchhod (1983) and Cattell (1984), the notion of support-verb construction includes constructions like *be angry* or *get loose*, where the support verb is *be* or one of its variants (Meyers 2007: 123). With this view, phrases like *be angry* or *be a genius* become support-verb constructions and therefore MWEs. Another consequence is that, in a support-verb construction, the core of the predicate may be an adjective or even a prepositional phrase (e.g. *be on time*) instead of a noun. Few computational linguists are familiar with these two ideas. But analysing all these expressions as support-verb constructions is consistent. They undergo semantic and syntactic phenomena observed with other support verb constructions:

(i) Syntactic operations produce constructs where the core of the predicate occurs without the verb, with the same meaning. For example, in the same way as *have* disappears in the alternance between *the habit the customer had* and *the customer's habit*, the verb *be* also disappears between *a customer who was angry* and *an angry customer*.

(ii) Other verbs can replace *be*, causing an aspectual or stylistic effect: compare *The customer was angry* with *The customer got angry*. This pair is parallel to *The customer had a habit / The customer gained a habit*.

(iii) There exist constructs with an additional causative or agentive subject and another verb,

[42] The appropriate definition of each PoS in this context is based on its possible syntactic contexts in sentences (cf. §5.1, footnote 40).

[43] Here, *free* means that the content of the slot, that is the noun phrase or the embedded clause, is variable.

as in (37) or in *The team was confident / Football made the team confident*. Such pairs are parallel to *The team had a goal / Football gave the team a goal*.

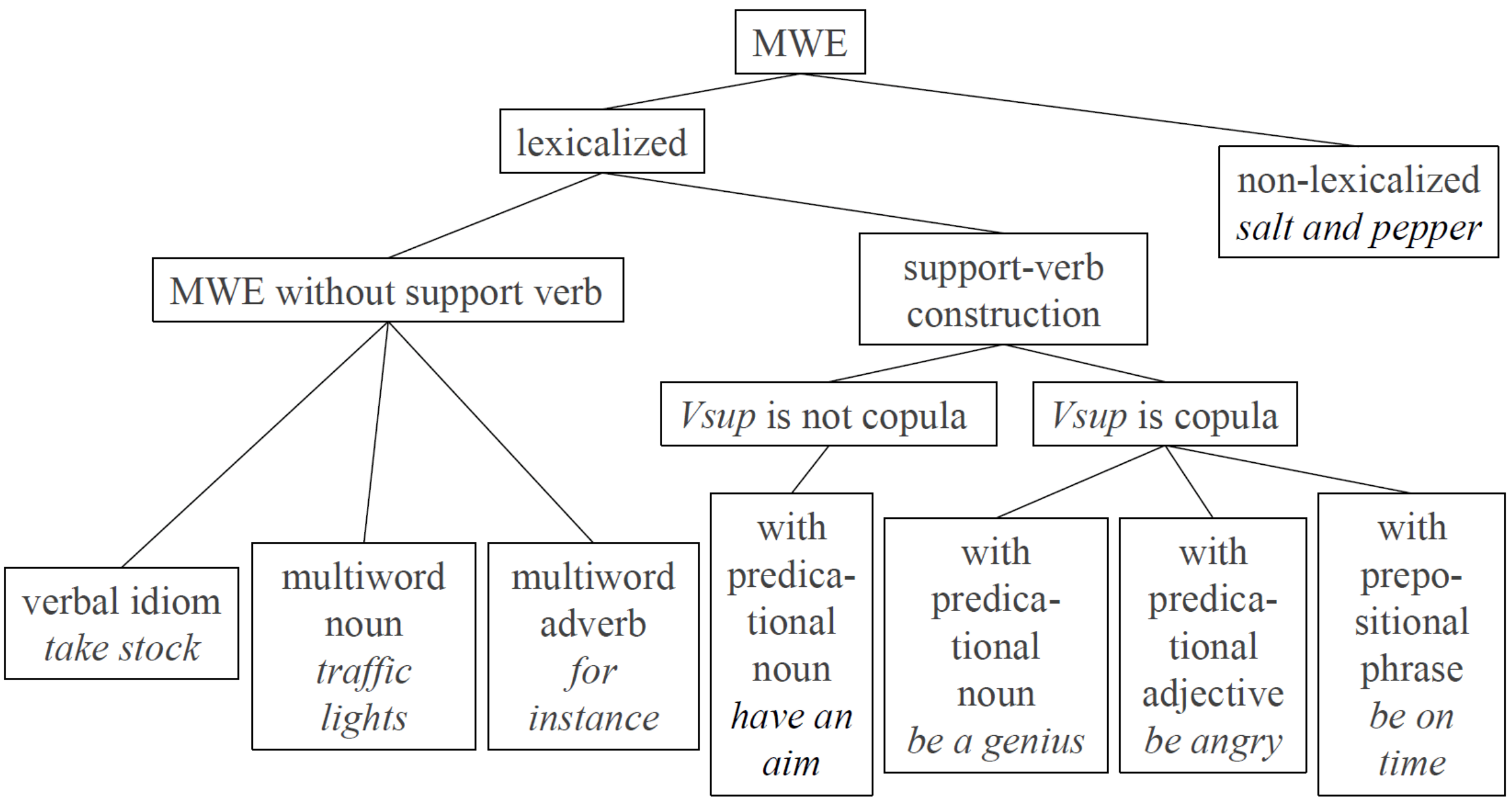

*Figure 2: Classification of MWEs where copula is considered a support verb.*

Figure 2 adopts this view and considers the copula (a linguistic term for *be* or its equivalent introducing a predicate) as a part of a support-verb construction. Prepositional phrases compatible with *be* shift to support-verb constructions. Since more expressions are considered MWEs than in Figure 1 and support-verb constructions become more diverse, they are divided in subclasses too, taking into account the PoS of the core of the predicate.[44] The core of the predicate may be either a word (*have an aim, be a genius, be angry*) or multiword (*have a point of view, be a smooth operator, be safe and sound, be on time*). There are two new categories: copulative constructions with a predicational adjective, for example *be angry, be safe and sound,*[45] and with a predicational noun, for example *be a genius*, *be a smooth operator*.

In Figure 2, the distinction between support-verb constructions or not is just above the decision about PoS. In addition, the PoS-based classification of support-verb constructions takes into account the PoS of the core of the predicate: noun (*aim, point of view, genius, smooth operator*), adjective (*angry, safe and sound*) or prepositional phrase (*on time*). This is an essential element of the diversity of these constructions. We can shift the PoS level of the decision tree above the support-verb-construction level, but then the tree will classify all support-verb constructions as verbal MWEs, and it will be desirable to add a second PoS level below, to take into account the PoS of the core of the predicate. Thus, the decision tree will have one more level than Figure 2 to define the same classes.

[44] The exact list of PoS under support-verb constructions and non-support-verb constructions depends on languages: in Arabic, Chinese or Korean, among others, predicational adjectives are used without a copula, and the class of copulative constructions with a predicational adjective is irrelevant for them.

[45] Non-predicational adjectives are those compulsorily attributive, for example *prime* in *This is John's prime role*: **This is John's role that is prime*. Another sense of *prime* corresponds to a predicational adjective: *21 is not prime*.

# 7 Conclusion

Something is to be learned from the experience of the last 20 years in respect of choosing features for classifying MWEs. Current practice routinely uses fuzzy features, or features defined in an imprecise way. On many occasions, a cluster of loosely correlated features is considered as a single feature. The choice of features with such flaws is likely to lead to classifications less fruitful for computational use. For example, describing the analysability/decomposability of verbal idioms is much less feasible and useful than describing their syntactic variation.

Selecting more appropriate features is not an easy task. It requires prioritizing good practices when studying MWEs. One of them consists in systematically assessing the reproducibility of observation of each feature, in order to obtain reliable repositories of lexical data. Another good practice is to check facts and predictions against the lexicon. It is understandable that some researchers try to avoid the patient examination of thousands of lexical entries for dozens of individual features, in the hope to reach the same results through other means. But it turns out that the laborious descriptive work they wish to elude is required not only to check hypotheses, but also to come across valid hypotheses: researchers that ignored large-coverage data constructed unverifiable hypotheses that won the attention of the community and resulted in loss of time.

The LG approach implements these good practices in descriptive and analytical work. On the basis of the results of such work carried out on several languages in parallel, I outlined an enhanced classification of MWEs.

# Acknowledgements

Thanks to the anonymous reviewers, to the scientific editors and to Alexis Neme, for their comments, suggestions and questions that considerably enriched this paper. Only I am responsible for its content.